\documentclass[letterpaper]{article} 
\usepackage{aaai2026}  
\usepackage{times}  
\usepackage{helvet}  
\usepackage{courier}  
\usepackage[hyphens]{url}  
\usepackage{graphicx} 
\usepackage{natbib}  
\usepackage{caption} 
\usepackage{amsmath}
\usepackage{amsthm}
\usepackage{amssymb}
\usepackage{booktabs}
\usepackage{multirow} 
\usepackage{algorithm}
\usepackage{algorithmic}

\usepackage{newfloat}
\usepackage{listings}
\DeclareCaptionStyle{ruled}{labelfont=normalfont,labelsep=colon,strut=off} 
\floatstyle{ruled}
\newfloat{listing}{tb}{lst}{}
\floatname{listing}{Listing}
\title{Who Can We Trust? Scope-Aware Video Moment Retrieval with Multi-Agent Conflict}
\author{
    Chaochen Wu\textsuperscript{\rm 1}, Guan Luo\textsuperscript{\rm 2}, Meiyun Zuo\textsuperscript{\rm 1\thanks{Corresponding author}}, Zhitao Fan\textsuperscript{\rm 3}
}
\affiliations{
    \textsuperscript{\rm 1}School of Information, Renmin University of China\\
    \textsuperscript{\rm 2}Institute of Automation, Chinese Academy of Sciences\\
    \textsuperscript{\rm 3}Independent Researcher, West Windsor Township, NJ, 08550\\

}

\usepackage{bibentry}

\begin{document}

\maketitle

\begin{abstract}
Video moment retrieval uses a text query to locate a moment from a given untrimmed video reference. Locating corresponding video moments with text queries helps people interact with videos efficiently. Current solutions for this task have not considered conflict within location results from different models, so various models cannot integrate correctly to produce better results. This study introduces a reinforcement learning-based video moment retrieval model that can scan the whole video once to find the moment's boundary while producing its locational evidence. Moreover, we proposed a multi-agent system framework that can use evidential learning to resolve conflicts between agents' localization output. As a side product of observing and dealing with conflicts between agents, we can decide whether a query has no corresponding moment in a video (out-of-scope) without additional training, which is suitable for real-world applications. Extensive experiments on benchmark datasets show the effectiveness of our proposed methods compared with state-of-the-art approaches. Furthermore, the results of our study reveal that modeling competition and conflict of the multi-agent system is an effective way to improve RL performance in moment retrieval and show the new role of evidential learning in the multi-agent framework. 
\end{abstract}

%

\section{Introduction}
\begin{figure}[!t]
\centering
\includegraphics[width=0.9\columnwidth]{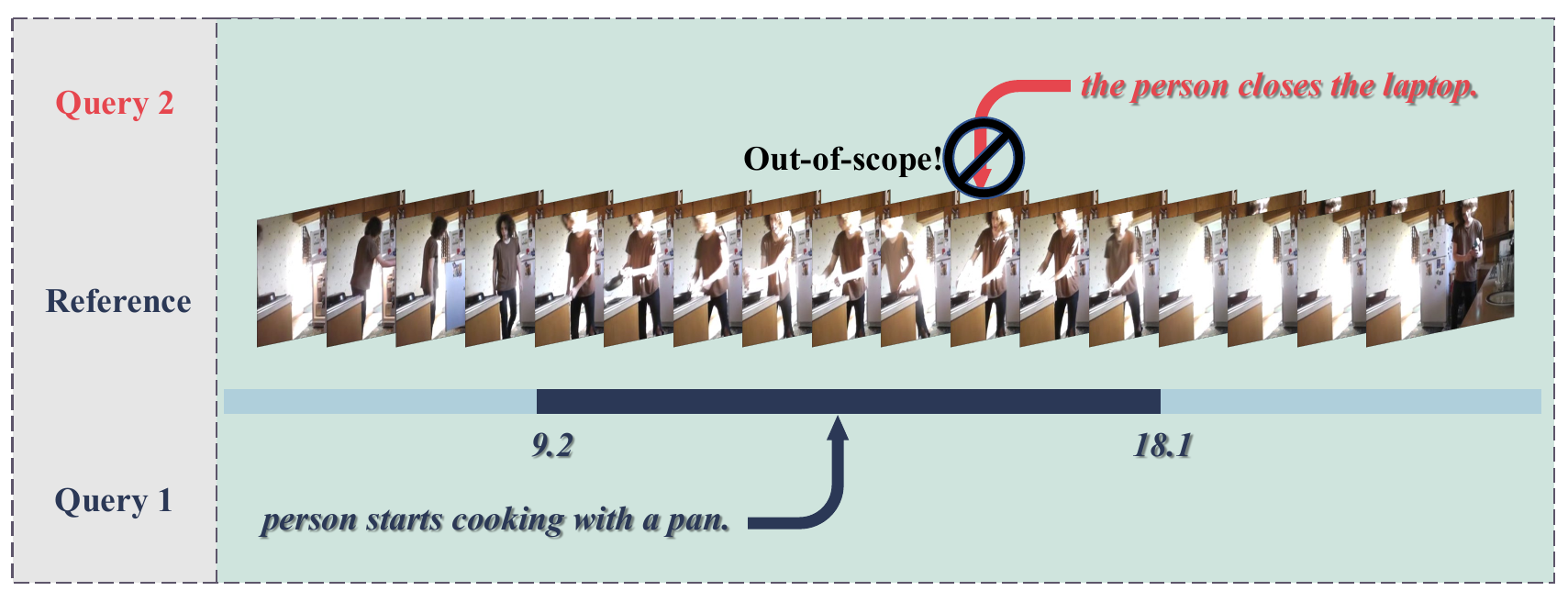}
\caption{An example of video moment retrieval. }
\label{fig1}
\end{figure}

In recent years, the development of visual recording technologies has created massive amounts of video data. However, it is time-consuming for people to locate moments they are interested in from long, untrimmed videos. Video moment retrieval(MR), also known as temporal grounding, uses text sentences to retrieve moments from videos, which brings convenience for searching information from untrimmed videos. Examples of applications of MR include finding desired scenes from movies, locating patients' dangerous accidents from hospital surveillance videos, and evaluating athletes' performance from sports game videos. Given a text query and a video reference, the cross-modal MR model should output a start timestamp and an end timestamp to indicate the corresponding moment in video reference, and we show an example in Figure \ref{fig1}. In the MR task, the MR model takes one query sentence and one video reference as input. For the query ``person starts cooking with a pan.",  the model outputs its corresponding timestamps (starting from 9.2 seconds and ending in 18.1 seconds). For the query ``the person closes the laptop.", the video reference does not have any corresponding moment, so it is an out-of-scope reference which does not have valid output.

One of the solutions for MR is a mode like the ranking task, which is also called anchor-based methods \cite{gao2017tall,hendricks2018localizing,liu2018cross,chen2019semantic,xu2019multilevel,zhang2019man}: First, it cut video reference in pre-defined window sizes and produced video segments; Then, it encoding video segments and query to features; Finally, it compute similarity between the query feature and video segments features and find the fittest video segments. Another type of solution called anchor-free approaches \cite{chen2018temporally,yuan2019find,lu2019debug,chen2020rethinking,zhang2020span,cao2021pursuit}, which using query features and video features to predict boundaries of video moments within all video frames. Most previous efforts only consider that videos are fully relevant to the query, but sometimes queries are not relevant to the video reference (out-of-scope). Some studies \cite{paul2021text,sun2021vsrnet,chen2023joint} study this issue with joint searching and grounding solutions. However, pursuing success in the search task requires the model to take extra training that needs training datasets for searching and may negatively affect the models' MR capacity. 

Besides the methods we already mentioned, some methods are built with reinforcement learning (RL). SM-RL \cite{wang2019language} used a recurrent neural network-based agent to observe a sequence of video frames dynamically and finally outputs the temporal boundaries of the given text query. Mimic a person watching the video, the RL agent can move the moment's boundary driven by rewards \cite{he2019read}, and in this study, the reward is computed by Intersection over Union (IoU) change: if an action increases IoU of the agent's current location and the objective location, the model will get a positive reward value. Otherwise, it will have a penalty. An improved method, MABAN \cite{sun2021maban}, uses multi-agent RL to retrieve video moment: it uses one agent to move the start boundary and one agent for the end,  and two agents share one observation network and environment. RewardTLG \cite{zeng2023rewardtlg} uses a vision-language pre-training model as a reward model, which provides flexible rewards different from previous RL methods.

\begin{figure*}[!ht]
\centering
\includegraphics[width=\textwidth]{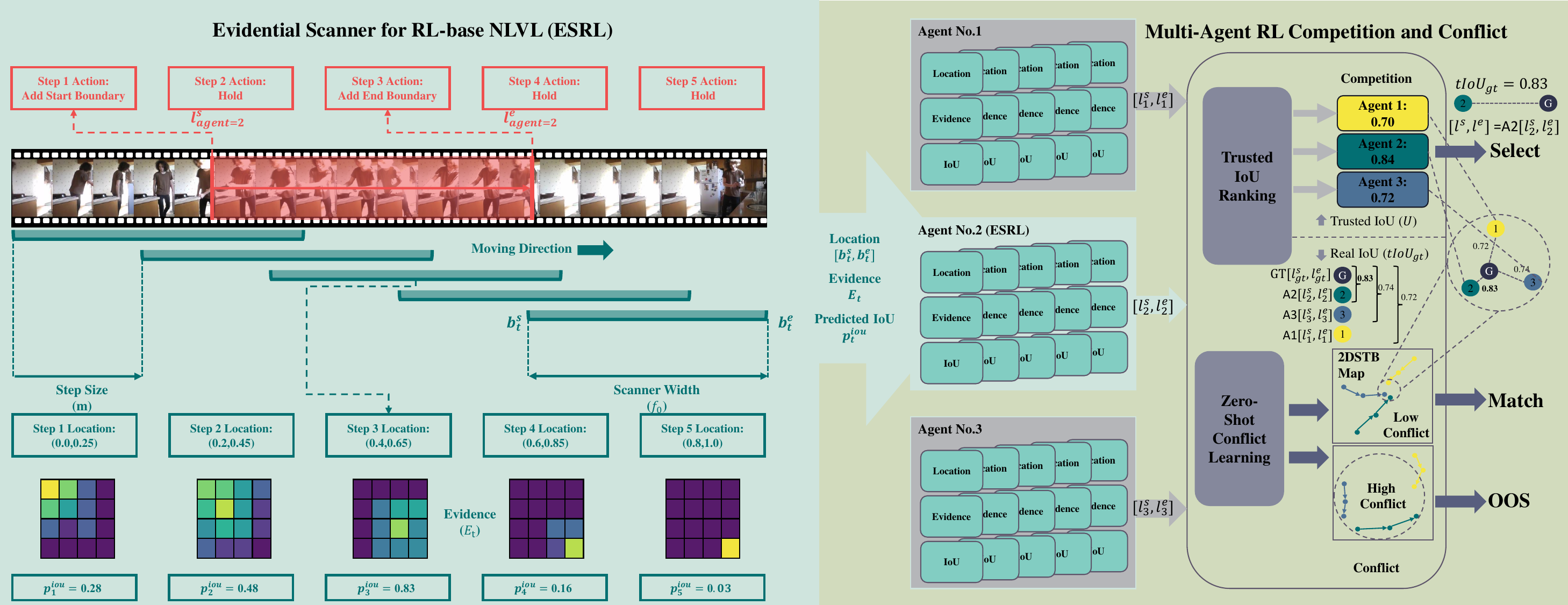}
\caption{Method overview. The left part is the action and output for ESRL. The right part is the architecture of MARLCC. ESRL produces IoU, location, and evidence for each step, and it moves in the right direction in a fixed window size and step size. For MARLCC, ESRL is one of the agents from three. MARLCC uses agents' trusted IoU to select the best result from \textit{competition}, and in the 2DSTB map, it is the ``closest" agent to the ground truth; MARLCC also can use \textit{conflict} to find OOS query, and in the 2DSTB map, the OOS query has higher \textit{conflict} than matched queries.}
\label{fig2}
\end{figure*}

Applying multi-agent cooperation to MR has these advantages: it decouples the start boundary movement with end boundary movement, which helps agents to find objective boundaries with higher flexibility; if one agent takes the wrong action and others take the correct actions, the final result may still approach to the objective moment's location. However, multi-agent cooperation has these limitations: the success of getting final moments on the basis of different agents collaborating properly, which brings extra complexity within the model design; the goal of cooperation restricts the designation of agents because we have to design agents' action types and reward functions carefully; if we want collaborated agents to produce an accurate result, each model have to perform well, but it is a challenging goal for black-box neural network models.

In this study, we propose a novel multi-agent system (MAS) framework for MR that sheds light on \textit{competition} and \textit{conflict} within agents. First, we incorporate evidential learning into RL-based MR models, and when RL agents take their actions, it will produce evidence and uncertainty for its relative boundary locations; second, inspired by previous MR methods that clip the video to segments, we build an RL-based MR model that scans all video feature with a fix window size and moves towards one direction; third, our method can produce agents' trusted IoU, which helps us to select the best localization output among different agents (\textit{competition}); fourth, various agents take their action independently, so we can evaluate the \textit{conflict} of their final locations to find out-of-scope (OOS) queries in a zero-shot manner. In real-world applications, users may give some queries that are beyond the scope of the video. What's more, OOS-aware MR models help people build video-level searching applications or select queries with corresponding moments from a query pool. Besides, our method does not require additional training processes like joint models \cite{chen2023joint,sun2021vsrnet}, so the MR ability will not have a chance to be weakened by the OOS detection task. 

The ultimate goal of this study is to propose a multi-agent RL method that has high performance for MR while detecting out-of-scope queries with a zero-shot method. Agents within MAS finish their jobs independently, enabling the designation of agents with higher flexibility. In general, we have two innovations in this study:
\begin{itemize}
    \item \textbf{Evidential Scanner for RL-base MR (ESRL)} Previous RL-based MR agents only move boundaries across part of the video \cite{he2019read,sun2021maban,zeng2023rewardtlg} and compute the IoU between current boundary with the ground truth. ESRL aims to scan all videos and predict IoU that covers the whole video. To detect an OOS query, scanning the whole video and seeing whether a matching moment exists may produce a more robust result. For the MAS designation, different agents have \textit{conflicts} between their final localization result, so ESRL applies evidential learning to produce evidence and uncertainty for the scanner's relative locations, and a trusted result should have stronger evidence for its location. 
    
    \item \textbf{Multi-Agent RL Competition and Conflict} Previously, few studies were interested in evidential learning in the MAS \cite{motlaghzadeh2020evidential,weng2020uncertainty}, and our studies firstly incorporate evidential MAS to MR. Different RL agents can produce location output with other agents orthogonally (independently), but their results conflict with each other. MARLCC takes agents' indicators like evidence to rank their trusted IoU, and we select the agent with the highest trusted IoU as the winner for \textit{competition}. For the first time, we found that for OOS queries, agents have an obviously higher \textit{conflict} within their moment outputs compared with queries with corresponding moments, prompting us to build a zero-shot method to detect OOS queries and apply it to video retrieval. 
\end{itemize}
The overview of our method is shown in Figure \ref{fig2}. We use the two-Dimensional Spatial-Temporal Boundary (2DSTB)  maps to visualize agents' boundary change and their movement in MR. 2DSTB map takes the coordinates to represent agents' start(x) and end(y) boundary positions, and the distance between two points inversely reflects their IoU. For each step, the agent adds a point representing its boundaries, and the line linking two dots is the agent's routine. The \textit{conflict} within agents is also clearly shown. 

\section{Method}
The MR task, which does not consider OOS queries, takes a video reference $V$ and a query sentence $Q$ as input. The output is the corresponding moment of the $Q$ in the form of start and end boundaries of the moment $[l^s,l^e]$. For OOS detection, it needs to output a label to indicate whether it is an OOS query, and if it is an OOS query, the moment is meaningless. Both ESRL and MARLCC produce a $[l^s,l^e]$ with a $V$ and a $Q$, and MARLCC can detect OOS without additional training. 

\subsection{Evdential learning for Agent's Locations and ESRL}
If agents take their action without cooperation, their MR contains \textit{conflicts}. Although a bunch of previous methods use IoU as their optimizing object, real IoU ($tIoU_{gt}$) computed by $[l^s,l^e]$ with the ground truth may not be as high as they predict ($p^{iou}$). Therefore, to find the best result, we need a trusted metric to decide the winner from agents' \textit{competition}.

\begin{figure*}[!t]
\centering
\includegraphics[width=0.9\textwidth]{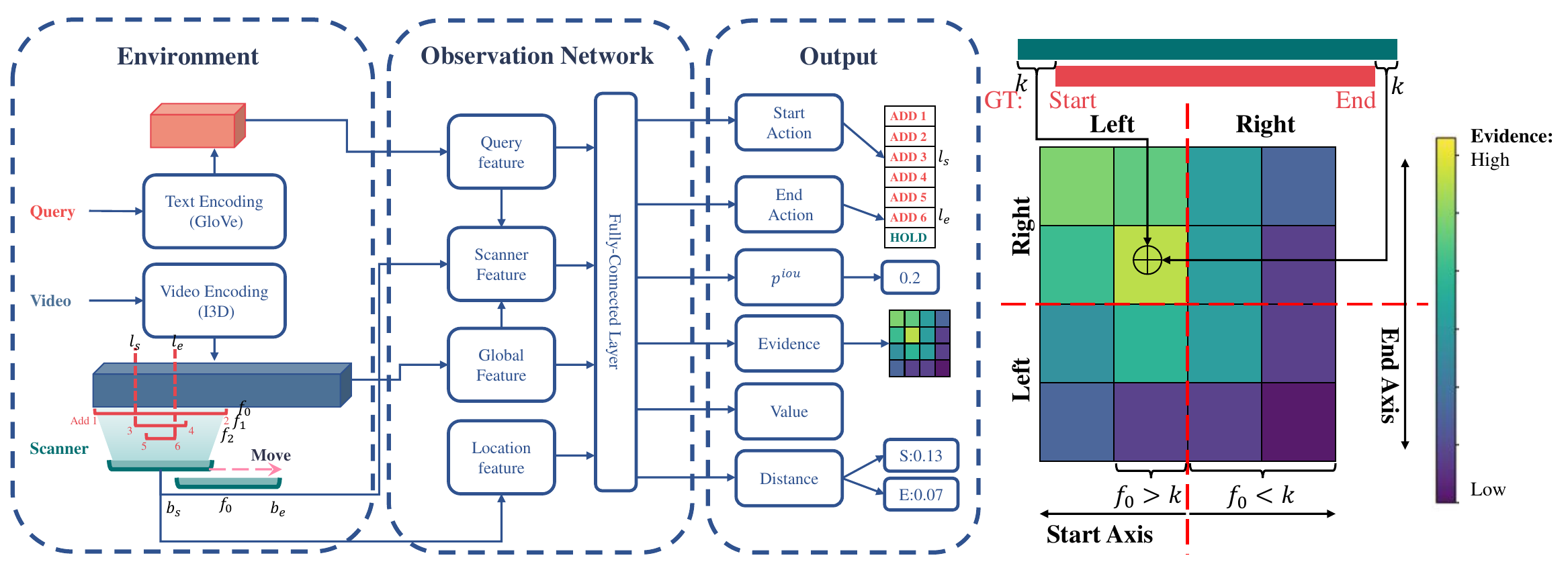}
\caption{ESRL architecture and two-dimensional representation of locational evidence. }
\label{fig3}
\end{figure*}

The scanner in ESRL refers to a fixed size ($f_0$) area that the model focuses on at the $t$th step, and its start and end boundaries are $[b^s_t, b^e_t]$.   Like previous RL-based MR agents, the ESRL model contains the environment and the observation network, and we present its architecture in Figure \ref{fig3}. ESRL uses the backbone as \cite{he2019read,sun2021maban} to build feature extract modules and the observation network. For getting video feature $V$, we used the I3D \cite{carreira2017quo}; for query encoding to get $Q$, we use GloVe \cite{pennington2014glove}; we use additional RNN-based encoding layer to process query and video feature. For collecting the scanner's feature $A_t$, we select video features included in the scanner's boundaries ($[b_t^s,b_t^e]$) and fuse it with query feature $Q$. For the location feature, we use a full-connected layer to encode $[b_t^s,b_t^e]$ to $L_t$. We concatenate $V$, $Q$, $A_t$, and $L_t$ together and use a fully connected layer to get the state vector in the $t$ step:
\begin{equation}
    O_t = FC_{O}(V \oplus Q \oplus A_t \oplus L_t)
\end{equation}
Where $\oplus$ is the concatenation operation, and $O_t$ is the environment feature in the $t$ step.

\textbf{Locational evidence} Evidential deep learning \cite{sensoy2018evidential,amini2020deep,chen2022dual} collects evidence for each class and builds a Dirichlet distribution of class probabilities over the collected class's evidence. \textit{Evidence} in here refers to a measure of the amount of support collected from input to support a sample to be classified to a class. 

In this study, we transform the boundary positions of the scanner to relative location classes, which allows us to integrate evidential deep learning into agent movement in reinforcement learning. For each step $t$, ESRL produces evidence $E_t$ for its relative location of the ground truth location. The class numbers $C = 16$ of relative location is the squared of the relative position number of one boundary (see the right part of Figure \ref{fig3}). One boundary has four possible relative positions: left ($k_t>f_0$), left ($f_0>k_t>0$), right ($f_0>k_t>0$), and right ($k_t>f_0$), where $f_0$ is the fixed window size of the scanner, and $k_t$ is the absolute distance for a current boundary to the ground truth boundary:
\begin{equation}
    k_t = |b_t - l_{gt}|
\end{equation}
For considering start and end boundaries together, the relative positions classes become $4 \times 4 = C$. We get evidence from $O_t$ as below: 
\begin{equation}
    E_t = FC_{evi}(O_t), E_t \in R^{1 \times C}
\end{equation}
Where $E_t = [e_1, \dots ,e_C]$. Evidential learning constructs a Dirichlet distribution with parameters $\alpha_j =e_j+1$ and $j=1 \dots C$ of class probability and generates evidence for each class label \cite{sensoy2018evidential}. The probability density function that is characterized by $C$ parameters $\alpha = [\alpha_1, \dots ,\alpha_C]$ of the Dirichlet distribution is:
\begin{equation}
    D(p|\alpha) = \begin{cases}
    \frac{1}{B(\alpha)}\prod_{j=1}^{C}p_j^{e_j} \quad p \in S_C\\
    0 \quad otherwise\\
    \end{cases}
\end{equation}
Where $p$ is a point on the C-dimensional unit simplex $S_C$, and $B(\alpha)$ is the $C$-dimensional multinomial beta function. Consider $D(p|\alpha)$ as the class probability distribution, and the optimization objective of evidential learning is:
\begin{equation}
    \mathcal{L}_{evi} = \sum_{j=1}^{C}y_j(\log(S) - \log(\alpha_j)), u = \frac{C}{S}
\end{equation}
Where $S=\sum_{j=1}^{C}\alpha_j$ is Dirichlet strength. $u$ is the uncertainty. From this formula, when the total evidence $S$ is low,  uncertainty $u$ will be high. Producing evidence and its derived uncertainty supports us in using evidence to indicate how certain an agent is aware of its relative location to the object moment and whether the agent's final result is worth trusting.

\textbf{Agent actions and loss computation} In previous RL methods, agents may not pass every place in the video feature and do not compute locational IoU and evidence of that place. ESRL uses a fixed window size ($f_0$) scanner to scan the video, and the step size ($m$) is also fixed, which is similar to previous anchor-base MR methods. The unique property of ESRL is it does not change the scanner's boundaries actively. Because window and step sizes are fixed, available actions of ESRL only contain two categories: ADD or HOLD. As the left part of Figure \ref{fig3}, the HOLD action means the agent does not change the start or end boundaries; six ADD actions for adding $[l^s,l^e]$ differ in the relative position within the scanner boundaries $[b^s,b^e]$.
\begin{equation}
    l^s_t = ADD_{s}([b^s_t,b^e_t]), b^s_t \leq l^s_t \leq b^e_t
\end{equation}
\begin{equation}
    l^e_t = ADD_{e}([b^s_t,b^e_t]), b^s_t \leq l^e_t \leq b^e_t
\end{equation}

Here, we briefly introduce how we perform training of ESRL and detail information in Supplementary Material. The reward function of ESRL is computed by the summation of start and end boundary rewards: 
\begin{equation}
    reward = r^s + r^e
\end{equation}
$r^s$ and $r^e$ are trained separately. The $r^s$ (or $r^e$ ) is computed by:
\begin{equation}
    r_t^s = \begin{cases}
        \rho, \quad if \quad a_t^s = HOLD \\
        \begin{cases}
            f_{dis}(\beta) ,\quad NOT \quad l^s_{t+1} \rtimes \Gamma_C  \\
            \begin{cases}
            f_{dis}(1 - d^s_{t+1}) , if \quad  d^s_{t+1} > d^s_{t} \\
            f_{dis}(-1 - d^s_{t+1}), if \quad  d^s_{t+1} \leq d^s_{t} \\
        \end{cases}
        \end{cases}
        
    \end{cases}
\end{equation}
\begin{equation}
f_{dis}(x) = x + d^s_{t} - \theta * d^s_{t+1}
\end{equation}
Where $a_t^s$ is the action. $d_t$ is absolute distance of $l_t$ with the ground truth boundary. $\rho$, $\beta$, and $\gamma$ are constant. $l_t \rtimes \Gamma$ mean boundary $l_t$ need to meet the condition $\Gamma$:
\begin{equation}
    \Gamma = \begin{cases}
       l_t < l^e_t,l_t \geq 0 \quad l_t \; is \;  l^s_t \\
       l_t > l^s_t,l_t \leq length \quad l_t \; is \; l^e_t \\
    \end{cases}
\end{equation}

We follow the traditional Actor-Critic algorithm to optimize the model to maximize the expected advantage, and this method learns both a policy $\pi(a_t|O_t)$ and a state-value $s_t$ that is produced by:
\begin{equation}
    \pi(a_t|O_t),s_t = GRU(O_t)
\end{equation}
The Mente Carlo sampling is adopted to optimize the loss of the policy $\mathcal{L}_{policy}$; we use the mean square error to optimize value loss $\mathcal{L}_{value}$. The model uses $O_t$ to compute predictions of the scanner's tIoU ($p_{t}^{iou}$), location, and distance as previous methods \cite{he2019read,sun2021maban} and uses ground truth for optimization. tIoU between the scanner's boundary $[b^s,b^e]$ and the ground truth $[l^s_{gt},l^e_{gt}]$ is computed by:
\begin{equation}
    tIoU_{gt}^b = \frac{\min(l^e_{gt}, b^e) - \max(l^s_{gt}, b^s)}{\max(l^e_{gt}, b^e) - \min(l^s_{gt}, b^s)}
\end{equation}

During the training process, we combined them together. 

\begin{equation}
    \mathcal{L}_{agent} = \mathcal{L}_{evi} + \mathcal{L}_{iou}+ \mathcal{L}_{dis}+ \mathcal{L}_{loc}+ \mathcal{L}_{policy} + \mathcal{L}_{value}
\end{equation}

\textbf{Other Agents: E-MABAN and E-DARK}
We added the evidential learning mechanism to another RL-based MR model( \cite{sun2021maban}) for creating E-MABAN as the second agent. Currently, RL-based MR methods use the ``included" area to compute the local video feature, and we use the feature produced by the "excluded" area to replace the "included" feature in MABAN to create the third agent E-DARK. In our study, MARLCC contains three RL agents: ESRL, E-MABAN, and E-DARK.

\subsection{MARLCC}
For ESRL or other RL agents with evidential learning, we can simplify their expression to:
\begin{equation}
     Agent_n(V, Q, O_{t})
\end{equation}
For each step $t$, $n$th agent output $E_{t}$, $p_{t}^{iou}$ (predicted IoU), $[b^s_{t}, b^e_{t}]$. We use three feed-forward layers follow a GRU to encode $\{E_{t}\}_{t=1}^T$, $\{p_{t}^{iou}\}_{t=1}^T$, $\{(b^s_{t}, b^e_{t})\}_{t=1}^T$.
\begin{equation}
    \{\hat{E_{t}}\}_{t=1}^T = FFN_{evi}(\{E_{t}\}_{t=1}^T)
\end{equation}
\begin{equation}
    \{\hat{p_{t}^{iou}}\}_{t=1}^T = FFN_{iou}(\{p_{t}^{iou}\}_{t=1}^T)
\end{equation}
\begin{equation}
    \{\hat{b}\}_{t=1}^T = FFN_{loc}(\{(b^s_{t}, b^e_{t})\}_{t=1}^T)
\end{equation}
\begin{equation}
    \{\theta_t\}_{t=1}^T = GRU(\{\hat{E_{t}}\}_{t=1}^T \oplus \{\hat{p_{t}^{iou}}\}_{t=1}^T \oplus \{\hat{b}\}_{t=1}^T)
\end{equation}
We got the $U$(Trusted IoU) by fusing the last output of the GRU layer and moment output in the last step:  
\begin{equation}
    U_n = FFN_{Tr}(\theta_T \oplus FFN_{loc}(l^s_n, l^e_n))
\end{equation}
The training goal of MARLCC for one agent is $U_n$ with its real tIoU ($tIoU_{gt}^n$) between $[l^s_n, l^e_n]$ and the ground truth $[l^s_{gt}, l^e_{gt}]$:
\begin{equation}
    \mathcal{L}_{Trust} = (U_n -  tIoU_{gt}^n)^2
\end{equation}
The loss function for MARLCC is computed by:
\begin{equation}
    \mathcal{L}_{MARLCC} =\sum_{n=1}^N(\mathcal{L}_{Trust})  +  \sum_{n=1}^N(\mathcal{L}_{agent})
\end{equation}
We select the agent with the highest $U$ as the winner of the \textit{competition}, and use its result as $[l^s,l^e]$.

\subsection{OOS detection with Conflict}
The conflict between agents for their moment output $[l^s_1, l^e_1]$ and $[l^s_2, l^e_2]$ can be computed by $\ell_1$-norm:
\begin{equation}
    \Vert c_{1,2} \Vert_1  =  |l^s_1 - l^s_2| + |l^e_1 - l^e_2|
\end{equation}
In the 2DSTB map, $\Vert c_{1,2} \Vert_1$ is the Manhattan distance. We can use $\Vert c \Vert_1$ as the metric to find OOS queries because the training object of MR is making different agents' results to approach the ground truth $[l^s_{gt}, l^e_{gt}]$. Therefore, the training process of agent $1$ and agent $2$ could be represented by:
\begin{equation}
    { \underset{w_1}{{\arg\min} \, \Vert c_{1,gt} \Vert_1}  } ,  { \underset{w_2}{{\arg\min} \, \Vert c_{2,gt} \Vert_1}  }
\end{equation}
Where $w_1$ and $w_2$ are parameter weights of agents 1 and 2.  OOS samples do not have a training object for $\Vert c_{1,gt} \Vert_1$ and $\Vert c_{2,gt} \Vert_1$, so $\Vert c^{OOS}_{1,2} \Vert_1$ tend to be higher than queries with corresponding moment $\Vert c^{match}_{1,2} \Vert_1$. The decision formula of a MAS with $N$ agents is written as:
\begin{equation}
    \eta = \max(\tbinom{N}{2}\Vert c \Vert_1)
\end{equation}
\begin{equation}
    Detect=\begin{cases}
    \eta > h, OOS\\
    \eta \leq h, match\\
    \end{cases}
\end{equation}
Where $h$ is a threshold parameter, $ \eta$ can also directly serve as the ranking score for zero-shot video searching: the greater the video and query match, the smaller $\eta$ is.
\section{Experiments}
Experiments of this study attempt to answer these questions: 
\begin{itemize}
    \item \textit{The performance of MARLCC and ESRL on benchmark MR datasets, and their comparison with SOTA methods.}
    \item \textit{The effect of zero-shot OOS detection with MAS and visualize conflicts for OOS and matched samples.  } 
    \item \textit{The importance of evidential learning in multi-agent RL for MR and how agents' competition improves RL performance.} 
\end{itemize}
\subsection{Datasets and Metrics}
Experiments for the MR task are performed on two widely used benchmark datasets: \textbf{Charades-STA} \cite{gao2017tall} with 9,848 daily indoor activities videos, 12,408 annotated queries for training, and 3,720 for testing. \textbf{ActivityNet-Caption} \cite{krishna2017dense}, which contains 19,209 videos and 71,957 query annotations.

Following prior works, the evaluation metrics we used for MR are $Acc@0.5$ and $Acc@0.7$, which is defined as the percentage of selected moments having tIoU scores between truth moments larger than 0.5 and 0.7. For OOS detection, we apply the accuracy and F1 scores and treat it as a classification task. In addition, we follow standard retrieval tasks to evaluate MARLCC on zero-shot video searching: we report Recall at rank K (R@K), and we select $K=\{1,10,100\}$.
\subsection{Experiment setting}
We use Adam with an initial learning rate of 0,001 for training optimization. MARLCC and three different agents, ESRL, E-MABAN, and E-DARK, are optimized together. Consider video length as 1.0, the step size of ESRL is set to 0.1; the windows size $f_0$ of its scanner is 0.12;  six ADD actions within the scanner $[0.0,0.12]$ from left to right is $[0, 0.02, 0.04, 0.08, 0.1, 0.12]$. The $\rho$, $\beta$, and $\gamma$ are set to 0, -0.8, and 0.4 respectively. The total number of steps ($T$) is set to 10, which is the same as other agents. We only use the RGB video feature for experiments on two datasets. The parameter $h$ for OOS was adjusted with the validation dataset with 1000 samples. 
\begin{table*}[!ht]
  \centering
  \begin{tabular}{c|ccccc}
  \hline
     \multirow{2}{*}{Type} & \multirow{2}{*}{Method} & \multicolumn{2}{c}{Charades-STA}  & \multicolumn{2}{c}{ActivityNet-Captions}  \\
                                                     \cline{3-4} \cline{5-6}
                                                   &  & Acc@0.5 & Acc@0.7 & Acc@0.5 & Acc@0.7   \\ 
    \hline
    \multirow{3}{*}{Other} & DRN \cite{zeng2020dense} & 53.09 & 31.75 & 45.45 & 24.36 \\
                       & 2D-TAN \cite{zhang2020learning} & 39.70 & 23.31 & 44.51 & 26.54 \\
                       & VSLNet \cite{zhang2020span} & 47.31 & 30.19 & 43.22 & 26.16 \\
                       & FMR \cite{gao2021fast} & 55.01 & \textbf{33.74} & 45.00 & 26.85 \\
                       & MomentDiff \cite{li2024momentdiff} & \textbf{55.57} & 32.42 & \textbf{46.52} & \textbf{28.43} \\
    \hline
     \multirow{8}{*}{RL} & SM-RL \cite{wang2019language} & 24.36 & 11.17 & -- & -- \\
                        & TripNet \cite{hahn2019tripping} & 38.29 & 16.07 & 32.19 & 13.93 \\
                        & TSP-PRL \cite{wu2020tree} & 37.39 & 17.69 & 38.76 & -- \\
                        & RWM \cite{he2019read} & 36.70 & -- & 36.90 & -- \\
                        & MABAN \cite{sun2021maban} & 52.12 & 27.74 & 44.88 & 25.66 \\
                        & RewardTLG \cite{zeng2023rewardtlg} & 52.30 & -- & -- & -- \\  
                        & ESRL ( Ours) & 50.70 & 29.17 & 45.05 & 25.34   \\
                        & MARLCC ( Ours) & \textbf{55.22} & \textbf{31.48}  & \textbf{46.29} & \textbf{26.55}   \\
    \hline
  \end{tabular}
  \caption{MR results on Charades-STA and ActivityNet-Captions datasets. }
  \label{table-1}
\end{table*}
\begin{table}[!ht]
  \centering
  \begin{tabular}{ccc}
  \toprule
     Dataset & Accuracy & F1   \\
    \hline
    Charades-STA & 66.33 & 68.68   \\
    \midrule
    ActivityNet-Captions & 88.42 & 88.99   \\
    \bottomrule
  \end{tabular}
  \caption{Result of zero-shot OOS queries detection.}
  \label{table-2}
\end{table}

\subsection{MR}
 We compare MARLCC and ESRL with existing SOTA RL-based MR methods: SM-RL \cite{wang2019language}, RWM \cite{he2019read}, TripNet \cite{hahn2019tripping}, TSP-PRL \cite{wang2019language}, MABAN \cite{sun2021maban}, RewardTLG \cite{zeng2023rewardtlg}. Besides, baseline methods also include DRN \cite{zeng2020dense}, 2D-TAN \cite{zhang2020learning}, FMR \cite{gao2021fast}, MomentDiff \cite{li2024momentdiff}, VSLNet \cite{zhang2020span}.

As shown in Table \ref{table-1}, our MAS method MARLCC achieves the best result compared with other SOTA RL-based methods, which shows that our method breaks the limitation of RL with single agent or agent collaboration. The evidential multi-agent framework is a useful way to improve MR model performance, and it can support agents with different architectures to work together and play their strengths to process different data samples. The ESRL (single-agent) got good experiment results in two datasets, and its results outperform RL-based methods except for MABAN. Our methods also outperformed some SOTA methods that are not based on RL, like DRN, 2D-TAN, and VSLNet. Our methof also approach recent MR methods like MomentDiff.
\subsection{OOS Detection and Application}
Besides traditional MR tasks, MARLCC can detect OOS samples without additional training. In Table \ref{table-2}, we test the accuracy score and F1 score for zero-shot OOS detection with MARLCC (datasets setting in Supplementary Material). Our zero-shot OOS detection result shows that with agents' ``conflict," we can find OOS queries without training. Although using extra models to detect OOS queries or multi-task learning may perform better for the OOS detection task, it needs additional supervised training, and whether joint training negatively affects MR performance needs to be studied profoundly. The OOS prediction results for ActivityNet-Captions are higher than for Charades-STA. Considering ActivityNet-Captions is a larger dataset than Charades-STA, more training samples may cause higher ``conflict" among agents. We can apply OOS detection for zero-shot video retrieval, and we present this result in Appendix.

\subsection{Ablation Study}
\begin{table}[!hb]
  \centering
  \begin{tabular}{ccccc}
  \hline

     Evidence & $p^{iou}$ & $[b^s, b^e]$ & Acc@0.5 & Acc@0.7   \\
    \hline
    $\times$ & \checkmark & \checkmark & 52.93 & 28.28 \\
    \checkmark & $\times$ & \checkmark & 54.03 & 28.87 \\
    \checkmark & \checkmark & $\times$ & 54.06 & 29.27 \\
    \checkmark & \checkmark & \checkmark & 55.22 & 31.48   \\
    \hline
  \end{tabular}
  \caption{Different features for MARLCC on Charades-STA dataset.}
  \label{table-4}
\end{table}

\begin{table}[!hb]
  \centering
  \begin{tabular}{ccccc}
  \hline

     ESRL & E-MABAN & E-DARK & Acc@0.5 & Acc@0.7   \\
    \hline
    \checkmark & $\times$ & $\times$ & 50.70 & 28.90 \\
    $\times$ & \checkmark & $\times$ & 51.40 & 28.84 \\
    $\times$ & $\times$ & \checkmark & 51.37 & 27.20 \\
    \midrule
    \checkmark & \checkmark & $\times$ & 53.40 & 30.22 \\
    \checkmark & $\times$ & \checkmark & 53.41 & 29.54 \\
    $\times$ & \checkmark & \checkmark & 53.36 & 29.17 \\
    \midrule
    \checkmark & \checkmark & \checkmark & 55.22 & 31.48   \\
    \hline
  \end{tabular}
  \caption{Different agents performance on Charades-STA dataset.}
  \label{table-5}
\end{table}
\begin{figure*}[!ht]
\centering
\includegraphics[width=\textwidth]{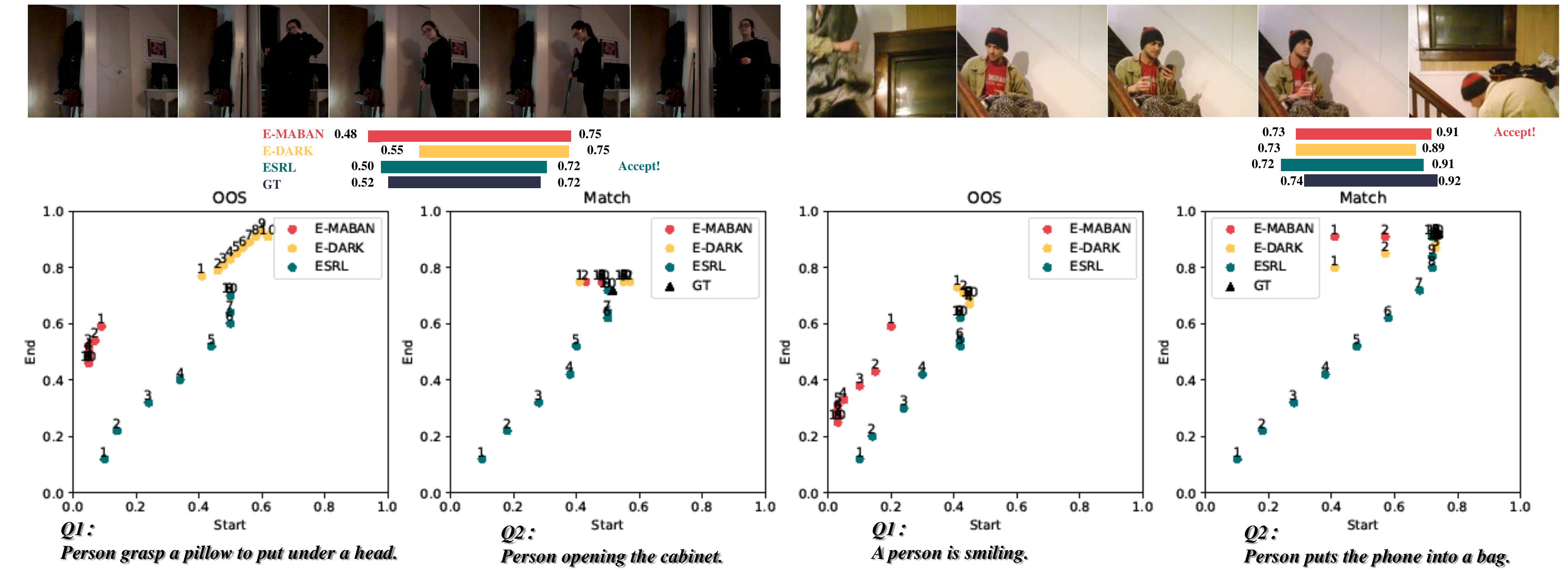}
\caption{MR examples with OOS queries and 2DSTB maps visualization for agents actions. }
\label{fig4}
\end{figure*}
Because MARLCC outperforms other RL-based MR models that use single agents or agents with collaboration, we want to explore how evidential learning contributes to MARLCC. Therefore, we disable evidence, $p^{iou}$, or $[b^s, b^e]$ from the MARLCC's input and run experiments to evaluate their MR performance; experiment results are shown in Table \ref{table-4}. Consistent with our estimation, evidential learning is irreplaceable for MARLCC. The MR results of the MARLCC model without evidence flop obviously to approach single agent's result, which shows that it cannot correctly deal with agents' \textit{competition}. Losing $p^{iou}$ or $[b^s, b^e]$ inputs has not caused an apparent drop in results of MARLCC, and their roles are not as crucial as evidence.

We are also interested in how different agent combinations affect MARLCC model performance, and we present the results of our method variants with one, two, and three agents in Table \ref{table-5}. From this table, we can conclude that our MAS-based methods, regardless of the agent numbers, outperformed all single-agent methods (ESRL, E-MABAN, E-DARK). MARLCC variants with two agents can also produce a better result than single-agent models. MARLCC, which has three agents, achieves the highest scores. Limited by time and computation resources, we did not run experiments for MARLCC that have four or more agents. Whether more agent numbers can produce a better result and the bottleneck of the performance improvement with agent numbers need to be further discussed. 
\subsection{MARLCC Visualization}
In Figure \ref{fig4}, We use two video examples and four queries to illustrate agents' \textit{conflict} and \textit{competition}, and we show how \textit{conflict} contributes to OOS detection with 2DSTB maps. Each video contains an OOS query and a matched query. For a matched query, we show different agents' results for the MR task and add the ground true location for comparison. As shown in the left video sample, ESRL got the highest IoU of its final moment localization results, and MARLCC accepted it as the selected output. As shown in the right video sample, E-MABAN got the highest IoU of its final results, and MARLCC takes E-MABAN's result as the final output. Three different agents(ESRL, E-MABAN, and E-DARK) produce three different outputs, and MARLCC uses evidence and other features to select the winner with the highest trusted IoU (\textit{competiton}).

We draw 2DSTB maps to visualize agents' movement and their final results. From the 2DSTB maps, the x-axis and y-axis indicate the start and end boundaries, respectively. The points and the number on the point show an agent's spatial location in the $t$th step ($[l_t^s,l_t^e]$). From the examples we draw, OOS samples show an obvious difference from matched samples. From the first query (OOS), the conflict is much higher than the second query (match), and ESRL, E-MABAN, and E-DRAK even move towards three different directions in the map. From the third query (OOS), ESRL and E-DARK have low \textit{conflict}, but E-MABAN and ESRL have a high \textit{conflict}. The fourth example clearly shows how different agents take their steps to converge on the ground truth point.
\section{Conclusion}
This study introduces an evidential multi-agent RL framework for MR (MARLCC) and a new RL agent for the MR task (ESRL). Our method helps improve RL methods prominently for MR tasks and shows its potential in OOS detection and video-level retrieval in a zero-shot way. Besides the new method we proposed, our experiments reveal a new phenomenon that agents' \textit{conflict} difference exists in OOS queries and matched queries. Limitations of our study include: first, we only discuss \textit{conflict} for final location result, and \textit{conflict} in agent movement is worth further exploring; second, MARLCC only considers RL-based methods, and studying how MARLCC integrates other MR architectures is an exciting issue.

\bibliography{aaai2026}

\begin{thebibliography}{33}
\providecommand{\natexlab}[1]{#1}

\bibitem[{Amini et~al.(2020)Amini, Schwarting, Soleimany, and Rus}]{amini2020deep}
Amini, A.; Schwarting, W.; Soleimany, A.; and Rus, D. 2020.
\newblock Deep evidential regression.
\newblock \emph{Advances in Neural Information Processing Systems}, 33: 14927--14937.

\bibitem[{Cao et~al.(2021)Cao, Chen, Shou, Zhang, and Zou}]{cao2021pursuit}
Cao, M.; Chen, L.; Shou, M.~Z.; Zhang, C.; and Zou, Y. 2021.
\newblock On Pursuit of Designing Multi-modal Transformer for Video Grounding.
\newblock In \emph{Proceedings of the 2021 Conference on Empirical Methods in Natural Language Processing}, 9810--9823.

\bibitem[{Carreira and Zisserman(2017)}]{carreira2017quo}
Carreira, J.; and Zisserman, A. 2017.
\newblock Quo vadis, action recognition? a new model and the kinetics dataset.
\newblock In \emph{proceedings of the IEEE Conference on Computer Vision and Pattern Recognition}, 6299--6308.

\bibitem[{Chen et~al.(2018)Chen, Chen, Ma, Jie, and Chua}]{chen2018temporally}
Chen, J.; Chen, X.; Ma, L.; Jie, Z.; and Chua, T.-S. 2018.
\newblock Temporally grounding natural sentence in video.
\newblock In \emph{Proceedings of the 2018 conference on empirical methods in natural language processing}, 162--171.

\bibitem[{Chen et~al.(2020)Chen, Lu, Tang, Xiao, Zhang, Tan, and Li}]{chen2020rethinking}
Chen, L.; Lu, C.; Tang, S.; Xiao, J.; Zhang, D.; Tan, C.; and Li, X. 2020.
\newblock Rethinking the bottom-up framework for query-based video localization.
\newblock In \emph{Proceedings of the AAAI Conference on Artificial Intelligence}, volume~34, 10551--10558.

\bibitem[{Chen et~al.(2022)Chen, Gao, Yang, and Xu}]{chen2022dual}
Chen, M.; Gao, J.; Yang, S.; and Xu, C. 2022.
\newblock Dual-evidential learning for weakly-supervised temporal action localization.
\newblock In \emph{European Conference on Computer Vision}, 192--208. Springer.

\bibitem[{Chen and Jiang(2019)}]{chen2019semantic}
Chen, S.; and Jiang, Y.-G. 2019.
\newblock Semantic proposal for activity localization in videos via sentence query.
\newblock In \emph{Proceedings of the AAAI Conference on Artificial Intelligence}, volume~33, 8199--8206.

\bibitem[{Chen et~al.(2023)Chen, Jiang, Xu, Cao, Mo, and Shen}]{chen2023joint}
Chen, Z.; Jiang, X.; Xu, X.; Cao, Z.; Mo, Y.; and Shen, H.~T. 2023.
\newblock Joint Searching and Grounding: Multi-Granularity Video Content Retrieval.
\newblock In \emph{Proceedings of the 31st ACM International Conference on Multimedia}, 975--983.

\bibitem[{Gao et~al.(2017)Gao, Sun, Yang, and Nevatia}]{gao2017tall}
Gao, J.; Sun, C.; Yang, Z.; and Nevatia, R. 2017.
\newblock Tall: Temporal activity localization via language query.
\newblock In \emph{Proceedings of the IEEE international conference on computer vision}, 5267--5275.

\bibitem[{Gao and Xu(2021)}]{gao2021fast}
Gao, J.; and Xu, C. 2021.
\newblock Fast video moment retrieval.
\newblock In \emph{Proceedings of the IEEE/CVF International Conference on Computer Vision}, 1523--1532.

\bibitem[{Hahn et~al.(2019)Hahn, Kadav, Rehg, and Graf}]{hahn2019tripping}
Hahn, M.; Kadav, A.; Rehg, J.~M.; and Graf, H.~P. 2019.
\newblock Tripping through time: Efficient localization of activities in videos.
\newblock \emph{arXiv preprint arXiv:1904.09936}.

\bibitem[{He et~al.(2019)He, Zhao, Huang, Li, Liu, and Wen}]{he2019read}
He, D.; Zhao, X.; Huang, J.; Li, F.; Liu, X.; and Wen, S. 2019.
\newblock Read, watch, and move: Reinforcement learning for temporally grounding natural language descriptions in videos.
\newblock In \emph{Proceedings of the AAAI Conference on Artificial Intelligence}, volume~33, 8393--8400.

\bibitem[{Hendricks et~al.(2018)Hendricks, Wang, Shechtman, Sivic, Darrell, and Russell}]{hendricks2018localizing}
Hendricks, L.~A.; Wang, O.; Shechtman, E.; Sivic, J.; Darrell, T.; and Russell, B. 2018.
\newblock Localizing Moments in Video with Temporal Language.
\newblock In \emph{Proceedings of the 2018 Conference on Empirical Methods in Natural Language Processing}, 1380--1390.

\bibitem[{Krishna et~al.(2017)Krishna, Hata, Ren, Fei-Fei, and Carlos~Niebles}]{krishna2017dense}
Krishna, R.; Hata, K.; Ren, F.; Fei-Fei, L.; and Carlos~Niebles, J. 2017.
\newblock Dense-captioning events in videos.
\newblock In \emph{Proceedings of the IEEE international conference on computer vision}, 706--715.

\bibitem[{Li et~al.(2024)Li, Xie, Xie, Zhao, Zhang, Zheng, Zhao, and Zhang}]{li2024momentdiff}
Li, P.; Xie, C.-W.; Xie, H.; Zhao, L.; Zhang, L.; Zheng, Y.; Zhao, D.; and Zhang, Y. 2024.
\newblock Momentdiff: Generative video moment retrieval from random to real.
\newblock \emph{Advances in neural information processing systems}, 36.

\bibitem[{Liu et~al.(2018)Liu, Wang, Nie, Tian, Chen, and Chua}]{liu2018cross}
Liu, M.; Wang, X.; Nie, L.; Tian, Q.; Chen, B.; and Chua, T.-S. 2018.
\newblock Cross-modal moment localization in videos.
\newblock In \emph{Proceedings of the 26th ACM international conference on Multimedia}, 843--851.

\bibitem[{Lu et~al.(2019)Lu, Chen, Tan, Li, and Xiao}]{lu2019debug}
Lu, C.; Chen, L.; Tan, C.; Li, X.; and Xiao, J. 2019.
\newblock Debug: A dense bottom-up grounding approach for natural language video localization.
\newblock In \emph{Proceedings of the 2019 Conference on Empirical Methods in Natural Language Processing and the 9th International Joint Conference on Natural Language Processing (EMNLP-IJCNLP)}, 5144--5153.

\bibitem[{Motlaghzadeh, Kerachian, and Tavvafi(2020)}]{motlaghzadeh2020evidential}
Motlaghzadeh, K.; Kerachian, R.; and Tavvafi, A. 2020.
\newblock An evidential reasoning-based leader-follower game for hierarchical multi-agent decision making under uncertainty.
\newblock \emph{Journal of Hydrology}, 591: 125294.

\bibitem[{Paul, Mithun, and Roy-Chowdhury(2021)}]{paul2021text}
Paul, S.; Mithun, N.~C.; and Roy-Chowdhury, A.~K. 2021.
\newblock Text-based localization of moments in a video corpus.
\newblock \emph{IEEE Transactions on Image Processing}, 30: 8886--8899.

\bibitem[{Pennington, Socher, and Manning(2014)}]{pennington2014glove}
Pennington, J.; Socher, R.; and Manning, C.~D. 2014.
\newblock Glove: Global vectors for word representation.
\newblock In \emph{Proceedings of the 2014 conference on empirical methods in natural language processing (EMNLP)}, 1532--1543.

\bibitem[{Sensoy, Kaplan, and Kandemir(2018)}]{sensoy2018evidential}
Sensoy, M.; Kaplan, L.; and Kandemir, M. 2018.
\newblock Evidential deep learning to quantify classification uncertainty.
\newblock \emph{Advances in neural information processing systems}, 31.

\bibitem[{Sun et~al.(2021)Sun, Long, He, Wen, and Lian}]{sun2021vsrnet}
Sun, X.; Long, X.; He, D.; Wen, S.; and Lian, Z. 2021.
\newblock VSRNet: End-to-end video segment retrieval with text query.
\newblock \emph{Pattern Recognition}, 119: 108027.

\bibitem[{Sun, Wang, and He(2021)}]{sun2021maban}
Sun, X.; Wang, H.; and He, B. 2021.
\newblock MABAN: Multi-agent boundary-aware network for natural language moment retrieval.
\newblock \emph{IEEE Transactions on Image Processing}, 30: 5589--5599.

\bibitem[{Wang, Huang, and Wang(2019)}]{wang2019language}
Wang, W.; Huang, Y.; and Wang, L. 2019.
\newblock Language-driven temporal activity localization: A semantic matching reinforcement learning model.
\newblock In \emph{Proceedings of the IEEE/CVF conference on computer vision and pattern recognition}, 334--343.

\bibitem[{Weng, Xiao, and Cao(2020)}]{weng2020uncertainty}
Weng, J.; Xiao, F.; and Cao, Z. 2020.
\newblock Uncertainty modelling in multi-agent information fusion systems.
\newblock In \emph{Proceedings of the 19th International Conference on Autonomous Agents and MultiAgent Systems}, 1494--1502.

\bibitem[{Wu et~al.(2020)Wu, Li, Liu, and Lin}]{wu2020tree}
Wu, J.; Li, G.; Liu, S.; and Lin, L. 2020.
\newblock Tree-structured policy based progressive reinforcement learning for temporally language grounding in video.
\newblock In \emph{Proceedings of the AAAI Conference on Artificial Intelligence}, volume~34, 12386--12393.

\bibitem[{Xu et~al.(2019)Xu, He, Plummer, Sigal, Sclaroff, and Saenko}]{xu2019multilevel}
Xu, H.; He, K.; Plummer, B.~A.; Sigal, L.; Sclaroff, S.; and Saenko, K. 2019.
\newblock Multilevel language and vision integration for text-to-clip retrieval.
\newblock In \emph{Proceedings of the AAAI Conference on Artificial Intelligence}, volume~33, 9062--9069.

\bibitem[{Yuan, Mei, and Zhu(2019)}]{yuan2019find}
Yuan, Y.; Mei, T.; and Zhu, W. 2019.
\newblock To find where you talk: Temporal sentence localization in video with attention based location regression.
\newblock In \emph{Proceedings of the AAAI Conference on Artificial Intelligence}, volume~33, 9159--9166.

\bibitem[{Zeng et~al.(2020)Zeng, Xu, Huang, Chen, Tan, and Gan}]{zeng2020dense}
Zeng, R.; Xu, H.; Huang, W.; Chen, P.; Tan, M.; and Gan, C. 2020.
\newblock Dense regression network for video grounding.
\newblock In \emph{Proceedings of the IEEE/CVF Conference on Computer Vision and Pattern Recognition}, 10287--10296.

\bibitem[{Zeng, Pan, and Han(2023)}]{zeng2023rewardtlg}
Zeng, Y.; Pan, K.; and Han, N. 2023.
\newblock RewardTLG: Learning to temporally language grounding from flexible reward.
\newblock In \emph{Proceedings of the 46th International ACM SIGIR Conference on Research and Development in Information Retrieval}, 2344--2348.

\bibitem[{Zhang et~al.(2019)Zhang, Dai, Wang, Wang, and Davis}]{zhang2019man}
Zhang, D.; Dai, X.; Wang, X.; Wang, Y.-F.; and Davis, L.~S. 2019.
\newblock Man: Moment alignment network for natural language moment retrieval via iterative graph adjustment.
\newblock In \emph{Proceedings of the IEEE/CVF Conference on Computer Vision and Pattern Recognition}, 1247--1257.

\bibitem[{Zhang et~al.(2020{\natexlab{a}})Zhang, Sun, Jing, and Zhou}]{zhang2020span}
Zhang, H.; Sun, A.; Jing, W.; and Zhou, J.~T. 2020{\natexlab{a}}.
\newblock Span-based Localizing Network for Natural Language Video Localization.
\newblock In \emph{Proceedings of the 58th Annual Meeting of the Association for Computational Linguistics}, 6543--6554.

\bibitem[{Zhang et~al.(2020{\natexlab{b}})Zhang, Peng, Fu, and Luo}]{zhang2020learning}
Zhang, S.; Peng, H.; Fu, J.; and Luo, J. 2020{\natexlab{b}}.
\newblock Learning 2d temporal adjacent networks for moment localization with natural language.
\newblock In \emph{Proceedings of the AAAI Conference on Artificial Intelligence}, volume~34, 12870--12877.

\end{thebibliography}

\end{document}